\documentclass{article}
\usepackage[T1]{fontenc}
\usepackage{iclr2027_conference,times}

\usepackage{amsmath,amssymb}
\usepackage{booktabs}
\usepackage{graphicx}
\usepackage{longtable}
\usepackage{microtype}
\usepackage{placeins}
\usepackage{float}
\usepackage{subcaption}
\usepackage{wrapfig}
\usepackage{xcolor}
\usepackage{hyperref}
\usepackage{url}
\usepackage{amsthm,mathtools}
\usepackage{array}
\usepackage{tabularx}
\usepackage{multirow}
\usepackage{enumitem}

\definecolor{flyteal}{HTML}{0D9488}
\definecolor{flynavy}{HTML}{17324D}
\newcommand{\method}{\textsc{FlyMark}}
\newcommand{\bitacc}{\mathrm{BitAcc}}
\newcommand{\sphere}{\mathbb{S}^{2}}
\newcommand{\va}{\mathbf{a}}
\newcommand{\vb}{\mathbf{b}}
\newcommand{\vc}{\mathbf{c}}
\newcommand{\vd}{\mathbf{d}}
\newcommand{\vo}{\mathbf{o}}
\newcommand{\vp}{\mathbf{p}}
\newcommand{\vq}{\mathbf{q}}
\newcommand{\vr}{\mathbf{r}}
\newcommand{\vs}{\mathbf{s}}
\newcommand{\vw}{\mathbf{w}}
\newcommand{\vx}{\mathbf{x}}
\newcommand{\mA}{\mathbf{A}}
\newcommand{\mR}{\mathbf{R}}
\newcommand{\mSigma}{\boldsymbol{\Sigma}}

\newcommand{\beeicon}{%
  \raisebox{-0.15em}{\includegraphics[height=1em]{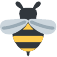}}%
}

\title{\beeicon\ \method: Training-Free Invisible Watermarking of 3D Gaussian Splatting via a Fruit Fly Connectome}

\author{%
Ziyuan Luo$^{1}$ \quad Haoliang Li$^{2}$ \quad Renjie Wan$^{1}$\thanks{Corresponding author.}\\
{\normalfont $^{1}$Department of Computer Science, Hong Kong Baptist University}\\
{\normalfont $^{2}$Department of Electrical Engineering, City University of Hong Kong}
}

\iclrfinalcopy
\hypersetup{
  pdftitle={FlyMark: Training-Free Invisible Watermarking of 3D Gaussian Splatting via a Fruit Fly Connectome},
  pdfauthor={Ziyuan Luo, Haoliang Li, Renjie Wan}
}
\begin{document}

\maketitle
\pagestyle{plain}

\begin{abstract}
A trained 3D Gaussian Splatting (3DGS) scene ships as a portable parameter array that can be
copied, pruned, requantized, or repackaged outside its training pipeline, so ownership evidence is
most useful when it lives in the released parameters and remains checkable long after the embedding
tooling is gone. Existing 3DGS watermarks typically tie embedding or extraction to scene
optimization, a learned decoder, or rendered views, so the evidence survives only as long as a
second trained artifact does. \method\ instead writes a keyed message into the parameters a 3DGS
file already stores. Its carrier directions are derived from the photoreceptors of a published
connectome, a citable versioned artifact that fixes the geometry exhaustively and leaves nothing to
tune per scene. A virtual observer reads cone-wise apparent luminance along a scene-normalized orbit
from stored centers, colors, and opacities; a keyed dithered quantization-index-modulation code
replicates each message bit across these observations; and one sparse bounded least-squares solve
realizes the targets through achromatic shifts of existing degree-zero colors under a hard
per-channel linear-RGB bound. All geometry and higher-order appearance parameters are preserved bit-identically, and extraction
needs only cone queries, rounding, and majority voting. Under a model-domain threat model on
synthetic and real scenes, \method\ attains high clean bit accuracy and visual fidelity while cleanly separating matched
from wrong keys.
\end{abstract}

\section{Introduction}
\label{sec:intro}

A fruit fly never chooses where to look. The directions its few thousand photoreceptors sample are
fixed by development: clustered, strikingly uneven, and essentially the same for every fly of the
species. What is new is that this pattern has now been written down, neuron by neuron, in a published
versioned connectome~\citep{berg2026malecns,nern2025visualconnectome}. A fly's field of view is now
an artifact anyone can cite. This paper asks what such an artifact is good for outside neuroscience,
and answers with an unlikely application: \textbf{watermarking 3D assets}. A watermark needs a place to look
that an owner today and an auditor years later can agree on exactly, and a \textbf{fly's retina} is such a
place.

We therefore \textbf{hire the fly} as a measuring instrument. Anatomy is an unusually good place to keep this
kind of agreement: the photoreceptor directions are public, reproducible from a versioned release,
tunable by no one, and densely clustered enough to hide thousands of symbols. \method\ turns this
virtual fly into a watermark for 3D Gaussian Splatting (3DGS), writing a keyed message into the parameters a
3DGS file already stores and reading it back by \textbf{flying an orbit around the scene}.

The need for such a witness is acute. 3DGS made radiance-field reconstruction real-time and
editable, but it ships as a plain array of anisotropic primitives~\citep{kerbl2023gaussians}, and
once a trained scene leaves its creator that array can be copied, pruned, requantized, or repackaged
entirely outside the original pipeline. Ownership evidence must therefore live in the released
parameters so that it travels with the file, survive the transformations asset pipelines routinely
apply, and leave rendered appearance unchanged under an auditable bound. A fourth requirement is
easy to overlook yet decisive: verification must succeed from the released file and a secret key
alone. Embedding and verification should be \textbf{training-free} after base-scene reconstruction
and \textbf{model-domain}, depending on no renderer convention and no artifact that must be kept
alive alongside the asset.

Published 3DGS watermarks trade off these requirements in different ways. GaussianMarker densifies
selected Gaussians and trains decoders~\citep{huang2024gaussianmarker}; GuardSplat learns
spherical-harmonic offsets together with an image decoder~\citep{chen2025guardsplat}; 3D-GSW
likewise relies on learned decoding~\citep{jang2025gsw}; GS-Hider couples secured
spherical-harmonic features with learned scene and message decoders~\citep{zhang2024gshider};
MarkSplatter uses splatter-image structure and a learned extractor~\citep{huang2025marksplatter};
and Splats in Splats hides 3D content in spherical-harmonic
coefficients~\citep{guo2026splats}. Learned extraction can improve image-domain robustness, but it
requires retaining a trained component, and image readouts inherit rendering conventions. This
leaves a complementary question: can a message be written into existing 3DGS parameters and
recovered from the released file by a deterministic procedure with no learned weights at all?

\begin{figure*}[t]
    \centering
    \includegraphics[width=\textwidth]{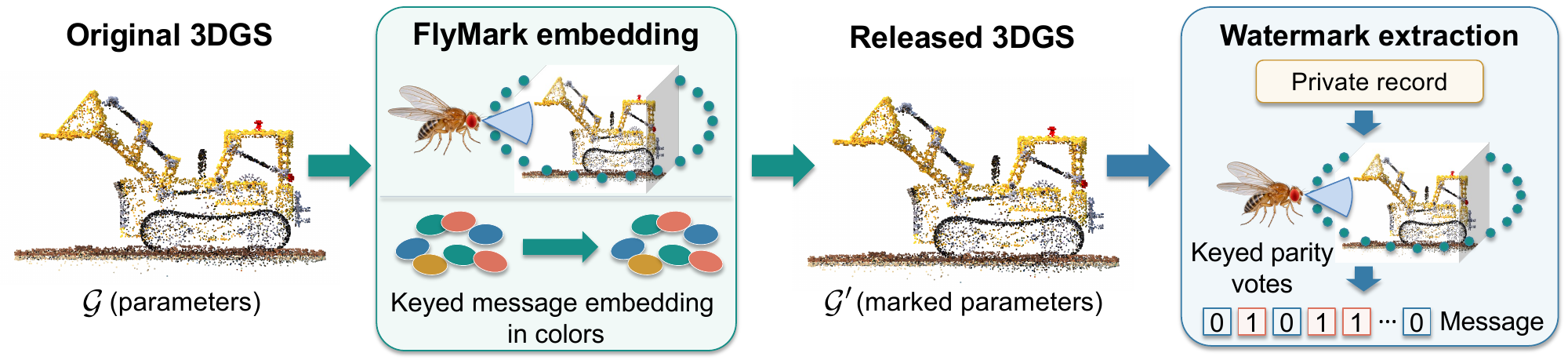}
    \vspace{-1.5em}
    \caption{{\method\ writes a message where a fly looks.} A virtual observer carries the
connectome's photoreceptor directions around the scene, and each direction reads an opacity-weighted
cone of Gaussians. Embedding perturbs only the stored degree-zero colors, so the marked render
$\mathcal{G}'$ is high visual fidelity from $\mathcal{G}$; extraction replays the same orbit
readout with the key. The fly, orbit, and bit glyphs are schematic.\vspace{-1.5em}}
    \label{fig:teaser}
\end{figure*}

Answering it takes more than perturbing parameters, because three obstacles constrain any such design.
\emph{First, there is no stable carrier index.} A symbol attached to one Gaussian disappears when that
primitive is pruned, and a row index is not a geometric identity after reordering, so a readout must be
addressed by scene-relative geometry and must aggregate many Gaussians at once. \emph{Second,
aggregation couples the edits.} Aggregated measurements share Gaussians, so moving one color perturbs
several measurements and writing the message becomes a coupled inverse problem rather than a sequence of
local edits. \emph{Third, a hard edit budget leaves some targets unattainable.} Under a strict
per-parameter fidelity bound and display-gamut limits, shared variables and active bounds guarantee that
part of the requested signal is never realized, so any code that needs every symbol to land exactly is
unusable.

\method\ answers these obstacles with three commitments. \textbf{Observe, do not index}: the carrier
is the fly's retina, projected into a common display plane and rotated with a virtual observer that
orbits the scene in scene-normalized units, so each of the $825$ directions defines a cone whose
opacity-weighted mean luminance is computed from stored centers, colors, and opacities alone.
\textbf{Code with redundancy and a key}: every message bit is replicated across the observation
array, and keyed dithered quantization index modulation~\citep{chen2001qim} turns each eligible
observation into a parity vote decoded by majority, so partially realized or locally damaged
observations are absorbed rather than fatal~\citep{cox1997spread}; following Kerckhoffs'
principle~\citep{kerckhoffs1883cryptographie,cayre2005watermarking}, the retina, the orbit, and the
decoder are public and secrecy resides only in the key-derived dither. \textbf{Solve once, under a
hard bound}: because luminance weights sum to one, an achromatic shift of a stored color moves every
observation containing it linearly, so all parity targets assemble into a single sparse bounded
least-squares solve~\citep{virtanen2020scipy} that rewrites only existing degree-zero colors within
a hard per-channel linear-RGB limit and the display gamut. Point count, positions, covariances,
opacities, and higher-order spherical harmonics are preserved bit-identically, and verification
repeats the orbit readout with cone queries, rounding, and voting (Figure~\ref{fig:teaser}). Our contributions can be summarized as folows:
\begin{itemize}[leftmargin=*,nosep]
    \item {the fly's eye as a watermark carrier}: we show that a published connectome's
    photoreceptor layout can serve as a public, non-tunable, reproducible measuring geometry for
    ownership evidence, read directly from existing 3DGS parameters;
    \item a keyed QIM code that spreads message bits over redundant cone observations and recovers
    them by deterministic parity voting and majority aggregation, with no learned decoder;
    \item a sparse linear color-realization formulation that fits all code targets in one
    gamut-aware bounded least-squares solve, modifying only existing colors under a hard
    per-channel shift bound.
\end{itemize}

With one configuration frozen before any held-out or real scene is evaluated, we measure clean bit
recovery, visual fidelity, robustness to model transformations, and wrong-key separation across
synthetic and real 3DGS scenes, against published baselines and matched parameter-space controls.
\method\ attains high clean bit accuracy at high fidelity, sustains recovery under model-domain
attacks, and cleanly separates matched from wrong keys, while a separate natural-image experiment
shows that the same retinal carrier and keyed QIM code support bounded watermarking of ordinary 2D
images.

\vspace{-0.5em}
\section{Related Work}
\vspace{-0.5em}
\label{sec:related}

\noindent\textbf{Radiance-field representations.}
NeRF introduced continuous view synthesis through coordinate networks~\citep{mildenhall2020nerf}. Anti-aliased and unbounded extensions~\citep{barron2021mipnerf,barron2022mipnerf360} and explicit factorizations that reduce or avoid multilayer-perceptron inference~\citep{fridovichkeil2022plenoxels,chen2022tensorf} broadened this representation family. 3DGS instead optimizes anisotropic Gaussians for differentiable splatting rather than ray marching~\citep{kerbl2023gaussians}; later work improves structure, anti-aliasing, compactness, and compression~\citep{lu2024scaffoldgs,yu2024mipsplatting,lee2024compactgs,niedermayr2024compressedgs}. These advances yield portable scene assets whose parameters can be distributed and transformed, motivating model-level watermarking.

\noindent\textbf{Watermarking radiance fields and 3DGS.}
NeRF watermarking uses joint optimization, image decoding, or codebooks~\citep{li2023steganerf,luo2023copyrnerf,jang2024waterf,luo2025nerfsignature}. 3DGS methods include uncertainty-guided densification and learned extraction (GaussianMarker~\citep{huang2024gaussianmarker}), secured spherical-harmonic features with learned decoders (GS-Hider~\citep{zhang2024gshider}), splatter-image representations (MarkSplatter~\citep{huang2025marksplatter}), learned robust decoding (3D-GSW~\citep{jang2025gsw}), learned spherical-harmonic offsets (GuardSplat~\citep{chen2025guardsplat}), and hidden 3D structure (Splats in Splats~\citep{guo2026splats}). Other work explores universal, generalizable, compressed, native, quantization-aware, policy-guided, and explainable watermarking~\citep{tan2024watergs,li2025gsmarker,in2026compmarkgs,qin2026ngsmarker,wang2026robust3dgsw,li2026w2m,cai2026wherewhatwhy}. \method\ instead uses existing degree-zero colors for a parameter-space code decoded without learned weights.

\noindent\textbf{Robust coding and watermark evaluation.}
Our keyed carrier draws on spread-spectrum redundancy~\citep{cox1997spread} and dithered-QIM reconstruction sets~\citep{chen2001qim}, placing repeated symbols in a scene-normalized orbit read from model parameters. Learned image watermarking studies distortion robustness~\citep{zhu2018hidden,tancik2020stegastamp,jia2021mbrs,zhang2019rivagan}, while neural-network ownership encodes signatures in model behavior or features~\citep{adi2018watermarking,rouhani2019deepsigns}. 3DGS removal attacks and broad-view protocols motivate measuring recovery alongside asset damage~\citep{huang2026gspure,zeng2026gmea,matsubara2026msvs}. We therefore evaluate bit accuracy, direct and ground-truth-referenced fidelity, attack curves, damage-matched robustness, and wrong-key calibration.

\section{FlyMark}
\label{sec:method}

\subsection{Embedding contract}

Let a static 3DGS asset be $\mathcal{G}=\{g_i\}_{i=1}^{N}$ with
$g_i=(\vx_i,\mSigma_i,\alpha_i,\vc_i^{(0)},\vc_i^{(>0)})$ collecting position, covariance, opacity, degree-zero color, and the remaining spherical-harmonic coefficients. Given a message $\vb\in\{0,1\}^{B}$ of length $B$ and a secret key $K$, embedding produces a marked asset $\mathcal{G}'$ that is verifiable from its parameters and satisfies the contract
\begin{equation}
N'=N,\quad
(\vx_i',\mSigma_i',\alpha_i',\vc_i^{(>0)'})=(\vx_i,\mSigma_i,\alpha_i,\vc_i^{(>0)}),\quad
\|\vr_i'-\vr_i\|_\infty\leq\epsilon,
\label{eq:contract}
\end{equation}
where $\vr_i\in[0,1]^3$ is the display-linear RGB value represented by $\vc_i^{(0)}$ and $\epsilon=0.02$. The embedding budget is thus a bounded achromatic edit of selected existing degree-zero colors, making the trade-off between message recovery and appearance explicit and auditable at the file level.

\subsection{Parameter-space readout}

\begin{figure*}[t]
    \centering
    \includegraphics[width=\textwidth]{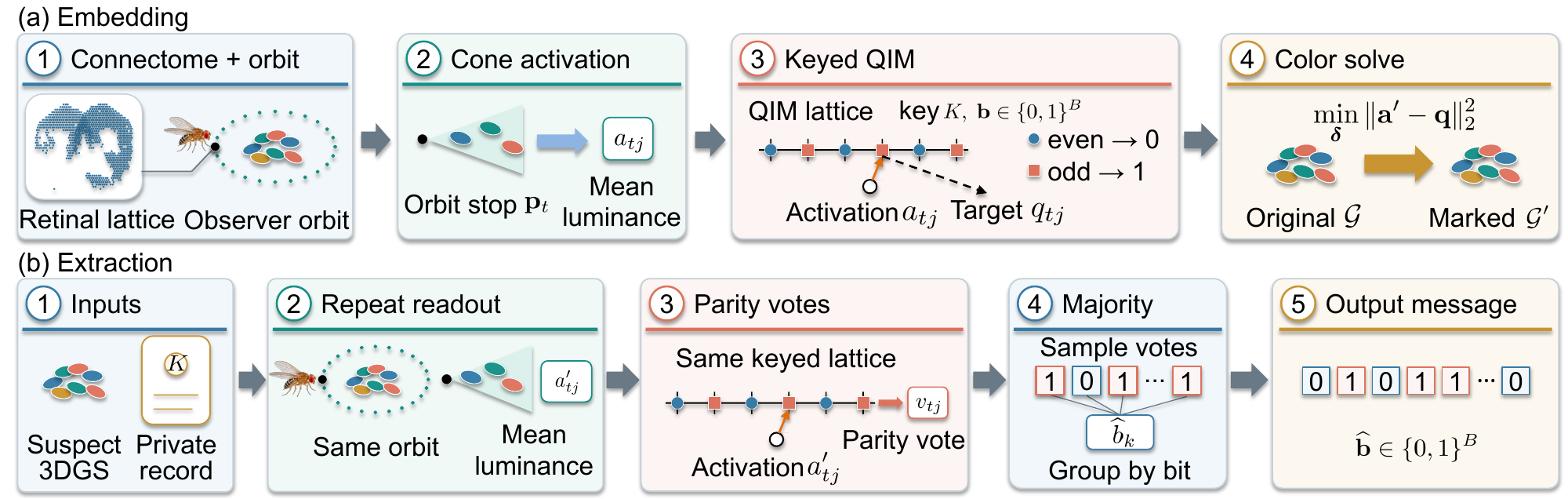}
    \caption{FlyMark's embedding and extraction paths. (a) The connectome-derived retinal lattice and observer orbit define a cone-wise mean-luminance readout $a_{tj}$ from the original Gaussian model $\mathcal{G}$. The key $K$ and message $\vb\in\{0,1\}^{B}$ determine keyed QIM targets $q_{tj}$: even lattice points encode 0 and odd points encode 1. A color solve adjusts the colors of existing Gaussians to produce the marked model $\mathcal{G}'$. (b) Given a suspect 3DGS model and the private record, the verifier repeats the same orbit readout to obtain $a'_{tj}$, converts each activation into a parity vote $v_{tj}$ on the keyed lattice, and takes a majority vote within each message-bit group to recover message. The displayed votes and message bits are schematic.\vspace{-1em}}
    \label{fig:method}
\end{figure*}

The detector measures an opacity-weighted apparent luminance
\begin{equation}
\ell_i=\alpha_i\,\vw^\top\vr_i,
\qquad \vw=(0.2126,0.7152,0.0722)^\top.
\label{eq:luminance}
\end{equation}
This functional uses three quantities that every 3DGS file stores explicitly: the Gaussian center, its degree-zero color, and its opacity. Reading them directly makes the carrier signal reproducible from the parameter array alone and stable across rasterizer implementations, tile orders, camera intrinsics, and image codecs, which is the property we need for a code that must be re-read after arbitrary repackaging. Figure~\ref{fig:method} summarizes the complete embedding and extraction paths and distinguishes the released marked asset from the private verification record.

\subsection{Connectome carrier and scene-normalized orbit}

A carrier geometry has to be agreed upon by the embedder and the verifier, reproducible years later, and rich enough to place thousands of redundant symbols. A published connectome satisfies all three at once: it is a citable artifact with a fixed sample list, it requires no design choices beyond a projection convention, and its density varies strongly across the visual field.

From the MaleCNS v1.0 reconstruction~\citep{berg2026malecns}, we obtain a two-dimensional display coordinate $(u,v)$ for each of 3,335 photoreceptors. Collapsing exact duplicate coordinates leaves $C=825$ distinct samples, and each sample becomes a pinhole direction
\begin{equation}
\vd_j=\frac{(u_j-\tfrac12,\,\tfrac12-v_j,\,1)}{\|(u_j-\tfrac12,\,\tfrac12-v_j,\,1)\|_2}\in\sphere.
\label{eq:lattice}
\end{equation}
The lattice therefore serves as a deterministic carrier geometry whose structure is inherited from anatomy; Appendix~\ref{app:ablation} visualizes the source samples and collapsed directions.

To express the readout in scene-independent units, we define a scene frame whose center $\vo$ is the midpoint of the Gaussian bounding box and whose radius $\rho$ is the 95th percentile of $\|\vx_i-\vo\|_2$. At $T=24$ equally spaced azimuths, a virtual observer occupies
\begin{equation}
\vp_t=\vo+1.5\rho
\begin{bmatrix}
\cos\eta\cos\theta_t & \sin\eta & \cos\eta\sin\theta_t
\end{bmatrix}^{\!\top},
\quad \eta=10^\circ,\quad \theta_t=2\pi t/T,
\label{eq:orbit}
\end{equation}
and looks toward $\vo$. For stop $t$ and carrier $j$, the set $\mathcal{C}_{tj}$ collects Gaussians whose center direction lies within angular radius $\gamma$ of the rotated carrier direction, where $\gamma$ is one half of the lattice's median nearest-neighbor angular spacing. The activation is the cone mean
\begin{equation}
a_{tj}=\frac{1}{|\mathcal{C}_{tj}|}\sum_{i\in\mathcal{C}_{tj}}\ell_i.
\label{eq:activation}
\end{equation}
An observation is eligible when its cone contains at least eight Gaussians, at least 80\% of them lie inside the 95th-percentile scene body, and $a_{tj}\in[0.05,0.95]$. These three criteria keep well-supported, scene-anchored, and unsaturated observations, which are the ones a bounded color edit can actually steer. Each message bit is repeated across the flattened $T\times C$ array, so a single disturbed cone is absorbed by its peers.

\subsection{Keyed dithered QIM}

Let $h(t,j)=(tC+j)\bmod B$ assign observation $(t,j)$ to message bit $b_{h(t,j)}$. From the private key $K$, a counter-based generator produces independent dithers $u_{tj}\in[0,\Delta)$ with step $\Delta=0.02$. The requested reconstruction point is the nearest lattice point of the desired parity,
\begin{equation}
q_{tj}=u_{tj}+\Delta\arg\min_{z\in\mathbb{Z}:\,z\bmod 2=b_{h(t,j)}}
\left|\frac{a_{tj}-u_{tj}}{\Delta}-z\right|.
\label{eq:qim}
\end{equation}
Keyed dithering randomizes the phase of the two parity cosets, and repeated assignment turns the large carrier array into a code of length $B$ with heavy replication. The construction follows classical QIM, moving each host signal onto a keyed quantization set~\citep{chen2001qim}.

\subsection{One bounded color-realization solve}

Perturbing a writable Gaussian by an achromatic shift $\delta_i$ gives $\vr_i'=\vr_i+\delta_i\mathbf{1}$. Since the luminance weights sum to one, Equation~\ref{eq:activation} responds linearly:
\begin{equation}
a'_{tj}=a_{tj}+\sum_{i\in\mathcal{C}_{tj}}\frac{\alpha_i}{|\mathcal{C}_{tj}|}\delta_i.
\label{eq:linear_activation}
\end{equation}
Stacking the eligible observations produces a sparse matrix $\mA$ and a residual $\vr=\vq-\va$, and we realize all targets at once through
\begin{equation}
\boldsymbol{\delta}^\star=\operatorname*{arg\,min}_{\boldsymbol{\ell}\leq\boldsymbol{\delta}\leq\boldsymbol{u}}\|\mA\boldsymbol{\delta}-\vr\|_2^2,
\qquad
l_i=\max(-\epsilon,-\min_c r_{ic}),\quad
u_i=\min(\epsilon,1-\max_c r_{ic}).
\label{eq:solve}
\end{equation}
The bounds enforce $\|\vr_i'-\vr_i\|_\infty\leq0.02$ and $\vr_i'\in[0,1]^3$ simultaneously. Gaussians whose stored degree-zero color already lies outside the display gamut are held out of the variable set and preserved bit-identically. One call to a bounded least-squares solver, using a trust-region reflective outer method with an LSMR inner solver, produces every shift~\citep{virtanen2020scipy}. Shared Gaussians and active bounds leave some carrier targets only partially realized, and redundant majority decoding absorbs those residuals, so embedding completes in a single solve.

\subsection{Extraction and ownership score}

The private verification record stores the key, the expected message, the carrier directions, the cone radius, and the embedding-time eligibility mask. Given a suspect model, the owner reconstructs the scene frame and orbit, evaluates Equations~\ref{eq:luminance} and~\ref{eq:activation}, and quantizes
\begin{equation}
\widetilde z_{tj}=\left\lfloor\frac{a'_{tj}-u_{tj}}{\Delta}+\tfrac12\right\rfloor,
\qquad v_{tj}=\widetilde z_{tj}\bmod 2 .
\label{eq:decode}
\end{equation}
For each index $k$, the decoded bit $\widehat b_k$ is the majority of the eligible votes with $h(t,j)=k$. The recovered message is $\widehat{\vb}=(\widehat b_0,\ldots,\widehat b_{B-1})$, and the primary recovery metric is
$\bitacc=B^{-1}\sum_k\mathbb{1}[\widehat b_k=b_k]$,
which is reported without any decision threshold. When an explicit ownership decision is required, the owner thresholds the number of correct bits at a preselected operating point, and Section~\ref{sec:security} calibrates that use separately.

\subsection{Implementation details}

The main evaluation sets $B=32$ message bits; the sensitivity study also evaluates $B=16$ and $B=64$. The implementation uses vectorized NumPy geometry and SciPy's sparse bounded least-squares solver~\citep{harris2020numpy,virtanen2020scipy}. The trust-region reflective outer solve uses tolerance $10^{-6}$ and at most 200 iterations, and its LSMR inner solves use tolerance $10^{-4}$ with the same iteration cap. Embedding stores the carrier, the key-derived protocol data, and the embedding-time eligibility mask. Fixed-mask decoding is the primary protocol, and recomputing eligibility from the received model provides a stricter secondary check. Extraction builds one spatial index per orbit stop and then performs cone queries, arithmetic, rounding, and majority voting, so both embedding and extraction run on CPU with the model file as their only input.

\section{Experiments}
\label{sec:experiments}

\subsection{Experimental Setup}

We protect a static vanilla 3DGS whose parameter array is available to the verifier. The owner holds a secret key and a verification record, while the lattice, the orbit construction, and the decoding algorithm may be published. An experimental cell is a scene/key pair: clean recovery is measured once per cell, and stochastic attacks additionally vary the attack seed. The adversary may serialize, reorder, quantize, prune, inject, perturb, smooth, or average the released asset. All methods start from the same scene reconstruction wherever an official baseline supports that scene.

\noindent\textbf{Datasets.}
We use eight NeRF Synthetic scenes~\citep{mildenhall2020nerf} and nine Mip-NeRF 360 scenes~\citep{barron2022mipnerf360}. Lego and Chair are development scenes; Ficus and Materials select the QIM step; Drums, Hotdog, Mic, and Ship are held-out Synthetic tests; all nine real scenes test external transfer. Each vanilla 3DGS base~\citep{kerbl2023gaussians} trains for 30,000 iterations from reconstruction seed 0. Keys 17, 29, and 43 yield 30 cells on the cross-method corpus and 27 cells across the nine real scenes for \method. The selected $\Delta=0.02$ and all other settings are frozen before held-out or real-scene evaluation; Appendix~\ref{app:protocol} details scene roles and selection.

\noindent\textbf{Baselines.}
We compare with three published 3DGS watermark systems and two matched parameter-space controls: 1) \textbf{GaussianMarker}~\citep{huang2024gaussianmarker}, uncertainty-guided Gaussian densification with a learned decoder; 2) \textbf{3D-GSW}~\citep{jang2025gsw}, a learned-decoder watermark; 3) \textbf{GuardSplat}~\citep{chen2025guardsplat}, which learns spherical-harmonic offsets and an image decoder; 4) \textbf{RandomCone-QIM}, which substitutes 825 blue-noise carrier directions while retaining \method's other stages; and 5) \textbf{SpatialHash-QIM}, which replaces cone aggregation with keyed spatial assignment under the same message length and color budget. Published methods use their official implementations; Appendix~\ref{app:protocol} specifies their configurations. Bit accuracy is normalized by each method's native message length. GuardSplat covers eight Synthetic scenes (24 cells); the others cover ten scenes (30 cells).

\noindent\textbf{Evaluation metrics.}
We measure message recovery, appearance, robustness, ownership separation, and cost. For \textit{message recovery}, the primary metric is bit accuracy, $\bitacc=\text{correct bits}/\text{message bits}$, and a threshold enters only at ownership operating points. For \textit{appearance}, direct fidelity compares marked against original renders with PSNR, SSIM~\citep{wang2004ssim}, and LPIPS using an AlexNet backbone~\citep{zhang2018lpips}, and ground-truth fidelity reports the change in novel-view PSNR against identical targets. For \textit{robustness}, we report bit-accuracy curves and an unweighted macro AUC over ordinal severity indices within each attack family. For \textit{ownership}, we measure wrong-key scores, false positives at a preselected cutoff, overwrite, and keyed-copy collusion. Structural audits compare serialized arrays for point count, geometry, opacity, non-DC spherical harmonics, gamut behavior, and the hard color bound.

\noindent\textbf{Attacks.}
The ten standard attacks include pruning, Gaussian injection, color, opacity and position noise, float16 round trips, color quantization, SH truncation, and row reordering. We additionally test targeted purification, local color smoothing, and spatial partial theft, plus overwrite and copy collusion for ownership analysis. Damage-matched comparison selects the most severe measured candidate inside a ground-truth PSNR-loss budget without consulting recovery or interpolating missing severities. Appendix~\ref{app:protocol} specifies severities and seeds.

\subsection{Results}
\label{sec:results}

\subsubsection{Clean Recovery and Fidelity}
\label{sec:clean}

\begin{table}[t]
    \caption{Clean message recovery and direct visual fidelity. Bit accuracy is the fraction of correctly recovered message bits, aggregated scene first. Fidelity compares marked renders against the corresponding unmarked 3DGS renders. GuardSplat covers the eight Synthetic scenes and the remaining methods cover all ten scenes. Arrows indicate the preferred direction.}
    \label{tab:main}
    \centering
    \small
    \resizebox{0.70\textwidth}{!}{\begin{tabular}{@{}lrrrrr@{}}
\toprule
Method & Cells & Bit acc. $\uparrow$ & PSNR $\uparrow$ & SSIM $\uparrow$ & LPIPS $\downarrow$ \\
\midrule
\textbf{FlyMark} & 30 & 99.27 & 44.21 & 0.993 & 0.0021 \\
RandomCone-QIM & 30 & 95.94 & 43.36 & 0.991 & 0.0026 \\
SpatialHash-QIM & 30 & 83.12 & 44.41 & 0.991 & 0.0022 \\
GaussianMarker & 30 & 98.75 & 30.11 & 0.954 & 0.0411 \\
3D-GSW & 30 & 86.67 & 32.59 & 0.970 & 0.0266 \\
GuardSplat & 24 & 62.11 & 41.06 & 0.996 & 0.0025 \\
\bottomrule
\end{tabular}}
\end{table}

Table~\ref{tab:main} places clean recovery beside marked-versus-unmarked visual fidelity. Across 30 scene/key cells, \method\ recovers 953 of 960 message bits (99.27\%) while reaching 44.21\,dB direct PSNR; recomputing eligibility from the received model still recovers 98.96\% of bits. Relative to ground truth, the mean novel-view PSNR change is 0.52\,dB, with a worst cell of 1.41\,dB.

The matched spatial-assignment control reaches comparable direct fidelity but lower recovery, supporting the value of redundant cone observations beyond the color budget alone. Among published systems, GaussianMarker comes closest in bit accuracy; \method\ leads the table while retaining higher direct fidelity and no learned decoder weights.

\begin{wraptable}{r}{0.52\linewidth}
    \vspace{-0.2em}
    \caption{Role-separated \method\ results. Held-out Synthetic and all nine real scenes are excluded from method selection.\vspace{-1em}}
    \label{tab:roles}
    \centering
    \small
    \resizebox{\linewidth}{!}{
    \begin{tabular}{@{}lrrrr@{}}
\toprule
Role & Cells & Bit acc. $\uparrow$ & PSNR $\uparrow$ & LPIPS $\downarrow$ \\
\midrule
Development & 6 & 100.00 & 43.82 & 0.0015 \\
Validation & 6 & 98.96 & 48.48 & 0.0003 \\
Held-out & 12 & 98.96 & 45.20 & 0.0021 \\
External real & 27 & 96.64 & 38.02 & 0.0045 \\
\bottomrule
\end{tabular}}
    \vspace{-\baselineskip}
\end{wraptable}

Table~\ref{tab:roles} tests the fixed carrier, orbit, and QIM settings beyond development scenes. Both the held-out Synthetic and nine-real-scene aggregates retain high clean bit accuracy and direct fidelity. This transfer spans white-background Synthetic scenes and varied real captures without retuning the carrier, orbit, or QIM step.

Appearance also holds away from the official trajectories. On the ten-scene cross-method corpus, multi-shell sampling of 37 to 158 viewpoints per scene, including cameras outside those paths, keeps direct PSNR above 38.7\,dB and LPIPS below 0.0060 on every scene, with per-scene values in Appendix~\ref{app:extended}. Structural audits of every clean cell confirm the embedding contract of Equation~\ref{eq:contract}, including identical point count, positions, scales, rotations, opacity, and non-DC spherical harmonics, the 0.02 linear-RGB bound on modified rows, and bit-identical preservation of pre-existing out-of-gamut rows.

\begin{figure*}[t]
    \centering
    \includegraphics[width=\textwidth]{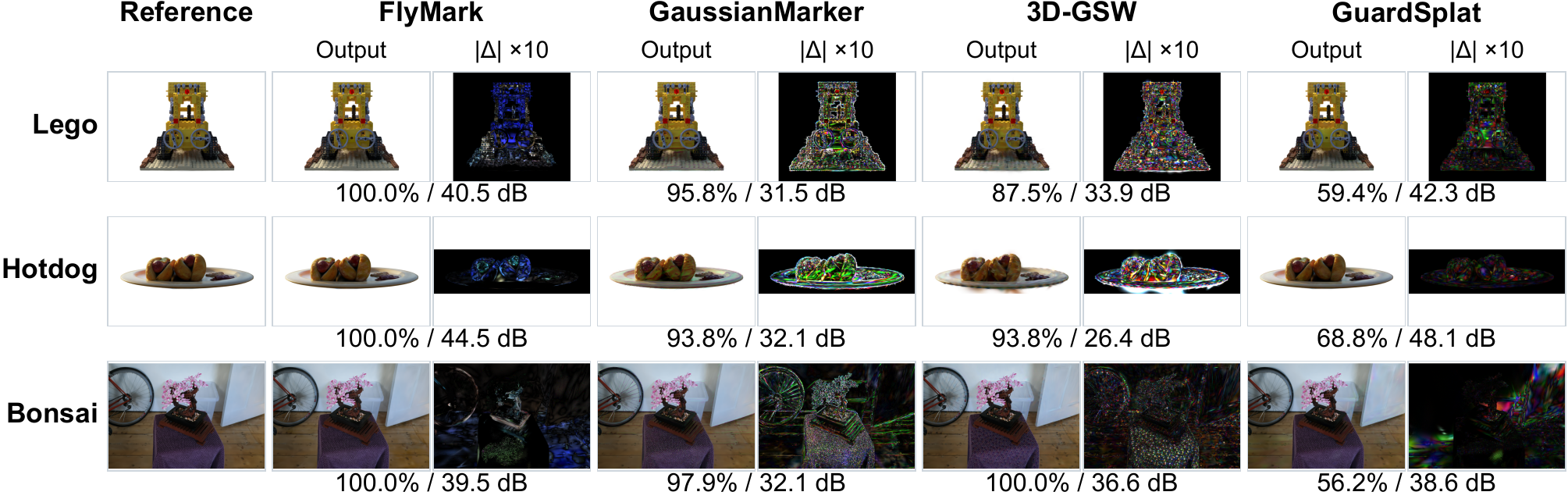}
    \caption{Registered-view qualitative comparison on Lego, Hotdog, and Bonsai. Every method is shown as an equal-size output/difference pair, where the right panel is the $10\times$ absolute RGB difference from the matched reference under a shared display transform. In the one-line annotation beneath each marked pair, the value before the slash is key-17 bit accuracy (\%) and the value after it is direct PSNR (dB) for the displayed frame. Each row uses identical cameras and crops across methods, and Bonsai uses the third registered view (frame 29). Tables~\ref{tab:main} and~\ref{tab:roles} report aggregate recovery and fidelity; Appendix~\ref{app:qualitative} shows all three registered views.\vspace{-1em}}
    \label{fig:qualitative}
\end{figure*}

The paired outputs and amplified differences in Figure~\ref{fig:qualitative} show the spatially diffuse color changes alongside intact message recovery in the displayed scenes.

\subsubsection{Robustness to Model Transformations}
\label{sec:robustness}

\begin{figure*}[t]
    \centering
    \includegraphics[width=\textwidth]{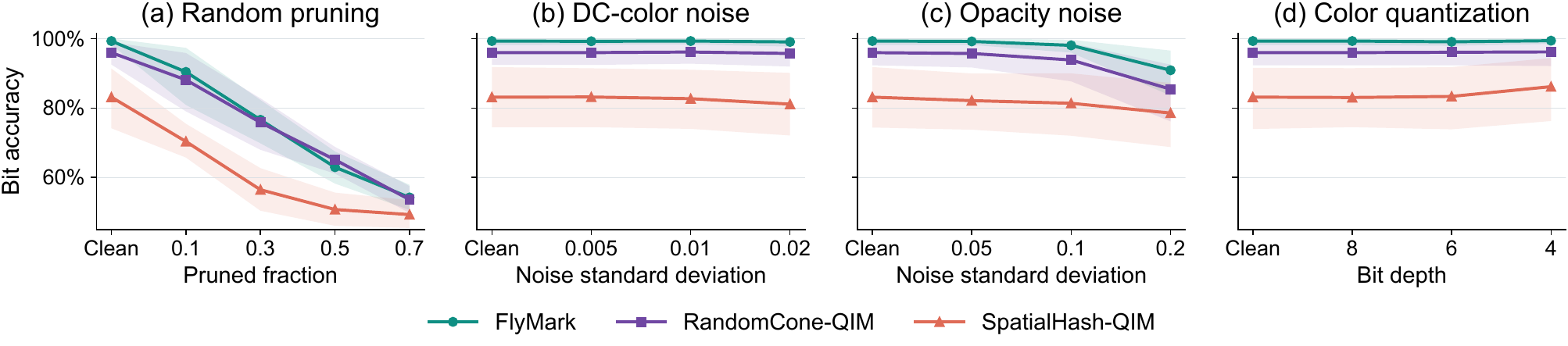}
    \caption{Bit accuracy under four representative model-attack families. Lines are scene-first means over ten scenes and three watermark keys, and shaded regions show hierarchical-bootstrap uncertainty. The clean point is repeated as severity zero.\vspace{-0.5em}}
    \label{fig:robustness}
\end{figure*}

We report model-attack recovery as per-severity curves (Figure~\ref{fig:robustness}) and ten-family AUC (Table~\ref{tab:all_attack_auc}).

\begin{figure}[t]
    \centering
        \includegraphics[width=0.94\columnwidth]{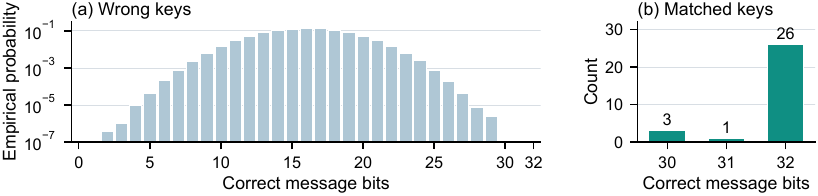}
    \caption{Wrong-key calibration. The left panel plots empirical probability by correct message bits over five million wrong-key trials (logarithmic vertical axis); the right panel counts clean matched-key scores across 30 scene/key cells. The preselected ownership cutoff is 30 correct bits. \vspace{-1em}}
    \label{fig:security}
\end{figure}

\begin{table*}[b]
\vspace{-0.5em}
    \caption{Bit-accuracy AUC for all ten standard attack families. Values average over ten scenes, three watermark keys, and three stochastic seeds where applicable. AUC integrates over ordinal severity within a family, and Macro is the unweighted family mean.}
    \label{tab:all_attack_auc}
    \centering
    \small
    \setlength{\tabcolsep}{2pt}
    \newcommand{\aucHeader}[1]{\begin{tabular}[c]{@{}c@{}}#1\end{tabular}}
    \resizebox{0.98\textwidth}{!}{\begin{tabular}{@{}l*{11}{c}@{}}
\toprule
\aucHeader{Method} & \aucHeader{Injection} & \aucHeader{DC\\noise} & \aucHeader{fp16} & \aucHeader{Quant.} & \aucHeader{Opacity\\noise} & \aucHeader{Opacity\\prune} & \aucHeader{Position\\noise} & \aucHeader{Random\\prune} & \aucHeader{Reorder} & \aucHeader{SH\\trunc.} & \aucHeader{Macro} \\
\midrule
FlyMark & 0.793 & 0.992 & 0.983 & 0.992 & 0.974 & 0.733 & 0.767 & 0.767 & 0.993 & 0.993 & 0.899 \\
RandomCone-QIM & 0.801 & 0.960 & 0.954 & 0.960 & 0.934 & 0.696 & 0.771 & 0.760 & 0.959 & 0.959 & 0.875 \\
SpatialHash-QIM & 0.539 & 0.827 & 0.718 & 0.837 & 0.814 & 0.544 & 0.535 & 0.609 & 0.831 & 0.831 & 0.709 \\
\bottomrule
\end{tabular}}
\end{table*}

Redundancy across directions and orbit stops sustains recovery through moderate deletion and parameter perturbation; noise or quantization near the QIM step $\Delta=0.02$ reduces it. Across the ten-family summary, \method\ has the strongest macro-AUC point estimate among the fully matched parameter-space constructions. Reordering, float16 round trips, SH truncation, and color requantization largely preserve the DC-color code, whereas deleting or displacing vote-carrying Gaussians is more disruptive. On the matched Lego, held-out Hotdog, and real Bonsai subset, its macro AUC is comparable to GaussianMarker and 3D-GSW; on the two Synthetic scenes supported by GuardSplat, it is higher.

Robustness is also assessed at comparable visual damage. Under a measured 2\,dB ground-truth PSNR-loss budget, \method\ retains 97.19\% of bits across all 30 available cells, similar to RandomCone-QIM and ahead of SpatialHash-QIM on commonly measured cells. Targeted attacks probe a different regime: local smoothing retains 97.92\% bit accuracy at severity 0.4, while opacity- and scale-guided purification retain 63.54\% and 70.83\% at severity 0.1, with respective ground-truth PSNR losses of 1.35 and 1.30\,dB. Appendix~\ref{app:attacks} gives the complete profiles, fidelity costs, and published-baseline coverage at the damage budget.

\vspace{-0.5em}
\subsubsection{Wrong-Key Calibration}
\vspace{-0.5em}
\label{sec:security}

Figure~\ref{fig:security} shows clear score separation: five million wrong-key trials, each changing the QIM dither and expected message, peak near 16 correct bits and never exceed 29, while matched scores range from 30 to 32. At the preselected 30-bit ownership cutoff, all matched cells and no wrong-key trial are accepted; unmarked-model, wrong-message, and cross-scene controls also yield no acceptance. This cutoff serves ownership decisions, while bit accuracy remains the unthresholded recovery metric. Appendix~\ref{app:attacks} examines overwrite and keyed-copy collusion.

\vspace{-0.5em}
\subsubsection{Component Attribution}
\vspace{-0.5em}
\label{sec:attribution}

\begin{table}[t]
    \caption{Component study on four development/validation scenes with key 17. Attacked bit accuracy averages component-attack conditions by scene; fidelity is direct marked-versus-original PSNR.}
    \label{tab:components}
    \centering
    \small
    \resizebox{0.88\textwidth}{!}{\begin{tabular}{@{}lrrrr@{}}
\toprule
Variant & Clean bit acc. & Recomputed & Attacked bit acc. & PSNR (dB) \\
\midrule
Full method & 100.00 & 99.22 & 68.91 & 46.02 \\
One orbit stop & 89.06 & 89.84 & 62.50 & 50.13 \\
No keyed dither & 91.41 & 89.84 & 66.41 & 46.29 \\
No scene-body gate & 99.22 & 99.22 & 68.75 & 45.83 \\
Differential cone pairs & 96.09 & 96.09 & 86.56 & 50.14 \\
Gamut only; no 0.02 bound & 100.00 & 100.00 & 78.44 & 30.62 \\
\bottomrule
\end{tabular}
\vspace{-2em}
}
\end{table}

\begin{table}[t]
    \caption{Implementation resource costs. \method\ uses CPU; published baselines use one GPU where applicable.}
    \label{tab:efficiency_detailed}
    \centering
    \small
    \begin{tabular}{@{}lrrrrr@{}}
\toprule
Method & {Embed median (s)} & {Extract/test (s)} & {Peak RSS (GiB)} & Params & {Decoder MB} \\
\midrule
FlyMark & 72.23 & 2.90 & 0.65 & 0 & 0.00 \\
GaussianMarker & 218.57 & 13.20 & 13.00 & 291,456 & 5.74 \\
3D-GSW & 1400.66 & 12.21 & 8.85 & 281,952 & 1.18 \\
GuardSplat & 226.46 & 13.59 & 1.94 & 205,344 & 1.22 \\
\bottomrule
\end{tabular}
\vspace{-1em}
\end{table}

Table~\ref{tab:components} isolates orbit redundancy, keyed dithering, eligibility, and the color constraint. One stop lowers clean bit accuracy by 10.94 points even as PSNR rises, so multi-stop voting contributes beyond stronger color edits. Removing dither costs 8.59 points; both variants have higher direct PSNR than the full method, separating their recovery losses from larger visual edits. Removing the scene-body gate changes clean and attacked means by less than one point. Differential cone pairs use a different detector and favor attacked recovery. Gamut-only writes retain clean bits but fall to 30.62\,dB, motivating the hard 0.02 bound.

One-factor sweeps show that a 64-bit message retains 96.48\% clean bit accuracy and twelve orbit stops retain 97.66\%. Varying the color bound shifts the fidelity and recovery trade-off without changing the carrier (Appendix~\ref{app:ablation}).

\vspace{-0.5em}
\subsubsection{Efficiency}
\vspace{-0.35em}
\label{sec:efficiency}

Table~\ref{tab:efficiency_detailed} reports a 72.23\,s median CPU embed, 2.90\,s extraction, and 0.65\,GiB median peak RSS. The one-time bounded solve writes the code; verification builds spatial indices and computes cone means and keyed votes without rendering or learned decoder weights. Its retained state is only the expected message, key, and carrier specification. A paired three-scene render check gives a mean marked/unmarked throughput ratio of 1.01, consistent with unchanged Gaussian count and geometry. Baseline timings use one GPU for embedding and the full test split for extraction.

\vspace{-0.5em}
\subsubsection{Extension to 2D Images}
\vspace{-0.35em}
For 2D images, we apply the retinal readout to bilinear luminance samples and realize keyed QIM with bounded pixel-color edits. On all 200 BSDS500 test images~\citep{arbelaez2011contour}, three keys each, saved 8-bit PNGs yield 99.95\% clean bit accuracy and 59.00\,dB mean direct sRGB PSNR. Appendix~\ref{app:2d} gives the protocol.

\vspace{-0.5em}
\section{Conclusion}
\vspace{-0.5em}
\label{sec:discussion}

\method\ embeds a keyed message in existing degree-zero colors of a static 3DGS asset. A connectome-derived carrier and orbit-based luminance readout define the code; one bounded color solve realizes it without new Gaussians, geometry changes, or decoder training. Across Synthetic and real scenes, the method combines accurate recovery and high visual fidelity; model-transformation and wrong-key tests assess robustness and ownership verification. These results establish biologically derived sampling as a practical model-domain watermark carrier.

\noindent\textbf{Limitation.} The present formulation targets static 3DGS assets. Future work can adapt its carrier readout and bounded embedding to NeRF scenes~\citep{mildenhall2020nerf} and VGGT-derived 3D reconstructions~\citep{wang2025vggt}, and extend it with temporally consistent readouts for 4D scenes.

\subsection*{AI use statement}
Generative AI
tools were not used to generate synthetic data sets, help develop theoretical models or conceptual
frameworks, formulate mathematical claims, propose or refine hypotheses, design or provide feedback
on research methodology or experiments, implement methods, clean or reformat datasets, or interpret
results; providing critical ingredients for proving mathematical claims, assisting in the writing of
proofs, translation, and qualitative or thematic data analysis are not applicable to this work.
Additionally, we used generative AI tools to edit the manuscript for readability and to search for
and identify relevant literature. We have reviewed all AI-assisted work: every AI-edited passage was
checked by the authors against the intended technical content, and every reference surfaced by an AI
tool was verified against the primary publication before citation. We take
responsibility for the final content of this work, including text, claims or artifacts produced with
the aid of generative AI.

\subsection*{Ethics statement}
This work studies ownership evidence for 3D assets using public datasets and a published connectome resource. The method infers no biological or personal attributes. As with other watermarks, ownership scores should be weighed alongside provenance records and legal context, and our wrong-key experiment provides an empirically calibrated operating point for that use.

\subsection*{Reproducibility statement}
Section~\ref{sec:method} specifies the complete readout, QIM rule, color bounds, and decoder. Appendix~\ref{app:protocol} details the evaluation protocol. The Supplementary Materials include code and a detailed README file for reproducing all reported results.

\bibliography{references}
\bibliographystyle{iclr2027_conference}
\clearpage

\appendix
\section{Appendix Overview}
\label{app:overview}

This appendix is organized as follows.
\begin{itemize}
    \item Section~\ref{app:protocol} specifies the datasets, baseline adaptations, metrics, statistics, and frozen hyperparameters.
    \item Section~\ref{app:derivation} expands the readout and constrained-realization derivation.
    \item Section~\ref{app:attacks} presents damage-matched, adaptive-attack, and ownership results.
    \item Section~\ref{app:extended} reports per-scene multi-shell fidelity.
    \item Section~\ref{app:ablation} covers carrier geometry and one-factor sensitivity.
    \item Section~\ref{app:2d} describes the bounded image-domain extension and its evaluation protocol.
    \item Section~\ref{app:qualitative} collects the registered qualitative views.
\end{itemize}

\section{Complete Experimental Protocol}
\label{app:protocol}

\subsection{Datasets and scene roles}

Table~\ref{tab:dataset_roles} gives the method-selection and test roles used throughout the paper.

\begin{table}[h]
    \caption{Dataset roles and clean evaluation cells. Every scene uses watermark keys 17, 29, and 43.}
    \label{tab:dataset_roles}
    \centering
    \small
    \resizebox{0.78\textwidth}{!}{\begin{tabular}{@{}llllr@{}}
        \toprule
        Dataset & Role & Scenes & Count & Cells \\
        \midrule
        NeRF Synthetic & Development & Lego, Chair & 2 & 6 \\
        NeRF Synthetic & Validation & Ficus, Materials & 2 & 6 \\
        NeRF Synthetic & Held-out test & Drums, Hotdog, Mic, Ship & 4 & 12 \\
        Mip-NeRF 360 & External real test & All nine scenes & 9 & 27 \\
        \bottomrule
    \end{tabular}}
\end{table}

NeRF Synthetic uses its canonical train/validation/test transforms and white background, and in particular Lego and held-out Hotdog each contain 100/100/200 views. Mip-NeRF 360 uses the official scene images and held-out rendering split with black background. Its nine scenes are Bicycle, Bonsai, Counter, Flowers, Garden, Kitchen, Room, Stump, and Treehill. Base models train for 30,000 iterations from reconstruction seed 0 with the public reference 3DGS implementation~\citep{kerbl2023gaussians}. Base training happens once per scene. On the ten-scene cross-method corpus, all compared methods start from the same saved model; the seven additional real scenes use the same frozen \method\ settings.

The QIM step is selected using Ficus and Materials only, from the registered candidate set $\{0.01,0.02,0.04,0.08\}$, after which $\Delta=0.02$ and all remaining settings are frozen before held-out Synthetic or external-real scenes are evaluated. The three secret keys apply to every main-method scene. Representative cross-method attack comparisons use key 17 and stochastic attack seed 101 on Lego, Hotdog, and Bonsai, which is the intersection for which comparable saved attack candidates exist across systems.

\subsection{FlyMark hyperparameters}

\begin{table}[h]
    \caption{Frozen \method\ configuration. The angular cone radius is derived from the carrier lattice instead of being specified as an absolute angle.}
    \label{tab:hyperparameters}
    \centering
    \small
    \resizebox{0.85\textwidth}{!}{\begin{tabular}{@{}ll@{}}
        \toprule
        Quantity & Value \\
        \midrule
        Message length $B$ & 32 bits \\
        Source carrier & 3,335 MaleCNS photoreceptors \\
        Distinct directions & 825 after exact UV collapse \\
        Orbit stops & 24 \\
        Orbit radius / elevation / phase & $1.5\rho$ / $10^\circ$ / $0^\circ$ \\
        Scene center / radius $\rho$ & bounding-box midpoint / 95th-percentile distance \\
        Cone radius & $0.5\times$ median nearest-neighbor angular spacing \\
        Minimum support / body fraction & 8 Gaussians / 0.8 \\
        Eligible activation interval & $[0.05,0.95]$ \\
        QIM step $\Delta$ & 0.02 \\
        Linear-RGB shift bound $\epsilon$ & 0.02 per channel \\
        Bounded solver & trust-region reflective + LSMR \\
        Outer tolerance / maximum iterations & $10^{-6}$ / 200 \\
        LSMR tolerance / maximum iterations & $10^{-4}$ / 200 \\
        Primary eligibility mode & fixed embedding-time mask \\
        Secondary eligibility mode & recomputed from received model \\
        \bottomrule
    \end{tabular}}
\end{table}

Table~\ref{tab:hyperparameters} records the exact frozen configuration used in the main study.

\subsection{Baselines and controls}

GaussianMarker uses its official uncertainty-aware embedding code, fixed 48-bit decoder checkpoint, and 2,000 embedding iterations~\citep{huang2024gaussianmarker}; we evaluate its native 48-bit message and normalize recovery by 48. 3D-GSW uses the official 32-bit decoder configuration, eight embedding epochs, and published loss weights~\citep{jang2025gsw}. GuardSplat uses its official 32-bit architecture with 200 decoder-training epochs and 100 watermark epochs~\citep{chen2025guardsplat}, and its aggregate evaluation covers the eight Synthetic scenes. For the complete qualitative comparison, we additionally run key 17 on Bonsai through GuardSplat's official LLFF camera path with the same 100-epoch configuration, and this targeted cell stays outside the aggregate baseline table. Seed adapters select deterministic messages while leaving baseline losses, renderers, camera sets, optimizers, and decoder architectures as published.

RandomCone-QIM uses 825 deterministic blue-noise directions over the same forward cap and otherwise shares \method's orbit, eligibility, QIM, and solve. SpatialHash-QIM keeps the message length and color budget while replacing cone aggregation with a keyed spatial assignment. The no-watermark control serializes an unmodified model through the same I/O path. Together these controls separate the contribution of the complete orbit code from that of a generic bounded perturbation or of file handling.

\subsection{Fidelity, attacks, and statistics}

Direct fidelity averages per-view metrics between marked and original renders over each official test split. Ground-truth quality evaluates both models against the same test targets and reports the PSNR change. PSNR and SSIM are computed in RGB, and LPIPS uses the AlexNet backbone. Structural audits compare saved model fields and require identical point count, positions, scales, rotations, opacity, and non-DC SH coefficients, a maximum absolute linear-RGB change of at most 0.02 on modified rows, no new or modified out-of-gamut rows, and bit-identical preservation of pre-existing out-of-gamut DC rows.

Standard attacks use these severities: random pruning $\{0.1,0.3,0.5,0.7\}$; opacity pruning $\{0.1,0.3,0.5\}$; Gaussian injection $\{0.05,0.1,0.3\}$; DC-noise standard deviation $\{0.005,0.01,0.02\}$ in linear RGB; relative opacity-noise standard deviation $\{0.05,0.1,0.2\}$; position-noise standard deviation $\{0.001,0.005,0.01\}\rho$; float16 round trip; 8/6/4-bit linear-RGB quantization; SH truncation to degree 2/1/0; and row reordering. Stochastic attacks use seeds 101, 202, and 303.

Each scalar metric is first averaged within a scene. Hierarchical-bootstrap intervals resample scenes and then cells within sampled scenes for 10,000 draws~\citep{efron1986bootstrap}. Paired contrasts resample or permute scene-level paired differences with Holm adjustment across the registered comparison family. Attack AUC prepends the clean point and integrates over equally spaced ordinal severity indices, and the macro value is an unweighted mean over attack families. Damage-matched selection takes the most severe \emph{measured} candidate whose ground-truth PSNR drop lies inside a given budget, decided without consulting recovery, and it performs no extrapolation across missing severities.

For efficiency measurements, \method\ runs on an Intel Xeon Platinum 8480C CPU; learned baselines use one NVIDIA H100 80GB GPU where applicable.

\section{Embedding and Extraction Details}
\label{app:derivation}

\subsection{Carrier membership}

At orbit stop $t$, let $\mR_t$ map the observer's local axes into world coordinates. A Gaussian with nonzero displacement from $\vp_t$ has direction
$\vs_{ti}=(\vx_i-\vp_t)/\|\vx_i-\vp_t\|_2$ and belongs to carrier $j$ when
\begin{equation}
\|\vs_{ti}-\mR_t\vd_j\|_2 \leq 2\sin(\gamma/2),
\end{equation}
which is the chord-distance form of an angular cone query. A separate scene-body indicator records whether $\|\vx_i-\vo\|_2$ falls below the 95th-percentile radius. That indicator drives the 80\% body-fraction eligibility gate, while activation means use every Gaussian center inside an eligible cone.

\subsection{QIM target and vote rule}

For normalized activation $y_{tj}=(a_{tj}-u_{tj})/\Delta$, the two reconstruction cosets are the even and the odd integers, so Equation~\ref{eq:qim} requests a change of at most $\Delta$ before shared-variable and gamut constraints take effect. Ties resolve toward the lower parity-constrained reconstruction point. At decoding time, rounding $y_{tj}$ to the nearest integer recovers its parity. For bit $k$, define
\begin{equation}
s_k=\sum_{(t,j):h(t,j)=k}m_{tj}(2v_{tj}-1),\qquad
\widehat b_k=\mathbb{1}[s_k>0],
\end{equation}
where $m_{tj}$ is the fixed or recomputed eligibility mask.

\subsection{Sparse linear system and color writeback}

Index the $M$ eligible observations by $m\leftrightarrow(t,j)$ and the $V$ writable Gaussians by $v\leftrightarrow i$. The system matrix is
\begin{equation}
A_{mv}=\begin{cases}
\alpha_i/|\mathcal{C}_{tj}|, & i\in\mathcal{C}_{tj},\\
0, & \text{otherwise}.
\end{cases}
\end{equation}
Because $\mathbf{1}^\top\vw=1$, this coefficient already contains the luminance response of an achromatic linear-RGB shift. A Gaussian may participate in many cones and stops, so Equation~\ref{eq:solve} reconciles overlapping target requests jointly. Bounds are built per Gaussian from $\epsilon$ together with the smallest channel-wise gamut headroom. Only the selected rows are converted back to DC spherical-harmonic coefficients, and untouched rows keep their original stored values. This selective writeback matters because a trained 3DGS may legitimately store raw degree-zero colors outside the display gamut even when its rendered color is clipped.

Embedding assembles the sparse matrix and calls the bounded solver once. Extraction performs no optimization: it builds one spatial index per orbit stop, queries the 825 cone directions, computes means, and majority-decodes, using only the model file and verification record.

\section{Additional Robustness and Ownership Results}
\label{app:attacks}

\subsection{Damage-matched standard attacks}

Table~\ref{tab:damage_matched} reports bit accuracy at a measured 2\,dB ground-truth PSNR-loss budget on the representative attack subset. For each condition, selection uses the most severe measured candidate inside the budget without consulting recovery. Cells without an eligible candidate remain missing rather than being extrapolated from more damaging severities.

\begin{table}[!thbp]
    \caption{Damage-matched bit accuracy at a measured 2\,dB ground-truth PSNR-loss budget on the representative Lego/Hotdog/Bonsai attack subset. Cells without a measured severity inside the budget are marked missing and are never extrapolated.}
    \label{tab:damage_matched}
    \centering
    \small
    \resizebox{0.60\textwidth}{!}{\begin{tabular}{@{}lrrr@{}}
\toprule
Method & Selected cells & Missing cells & Bit acc. \\
\midrule
FlyMark & 30 & 0 & 97.19 \\
RandomCone-QIM & 30 & 0 & 98.02 \\
SpatialHash-QIM & 20 & 10 & 82.97 \\
GaussianMarker & 0 & 30 & \textemdash{} \\
3D-GSW & 10 & 20 & 100.00 \\
GuardSplat & 1 & 19 & 56.25 \\
\bottomrule
\end{tabular}}
\end{table}

\subsection{Adaptive attacks}

\begin{table}[!thbp]
    \caption{Adaptive attacks on Lego, Hotdog, and Bonsai. Bit accuracy averages the available evaluation units, and ground-truth PSNR loss quantifies the accompanying asset change. Spatial partial theft includes three spatial selections per scene.}
    \label{tab:adaptive}
    \centering
    \small
    \resizebox{0.78\textwidth}{!}{\begin{tabular}{@{}llrrr@{}}
\toprule
Attack & Severity & Scenes/cells & Bit acc. & Mean GT drop (dB) \\
\midrule
Purification by opacity & 0.1 & 3 & 63.54 & 1.35 \\
Purification by opacity & 0.3 & 3 & 50.00 & 2.78 \\
Purification by opacity & 0.5 & 3 & 46.88 & 6.47 \\
Purification by scale & 0.1 & 3 & 70.83 & 1.30 \\
Purification by scale & 0.3 & 3 & 59.38 & 1.73 \\
Purification by scale & 0.5 & 3 & 57.29 & 3.63 \\
Local color smoothing & 0.02 & 3 & 100.00 & 1.36 \\
Local color smoothing & 0.05 & 3 & 100.00 & 1.50 \\
Local color smoothing & 0.1 & 3 & 100.00 & 1.82 \\
Local color smoothing & 0.2 & 3 & 100.00 & 2.59 \\
Local color smoothing & 0.4 & 3 & 97.92 & 4.20 \\
Spatial partial theft & 0.25 & 9 & 52.08 & 23.26 \\
Spatial partial theft & 0.5 & 9 & 50.00 & 20.21 \\
Spatial partial theft & 0.75 & 9 & 58.68 & 15.89 \\
\bottomrule
\end{tabular}}
\end{table}
\FloatBarrier

Table~\ref{tab:adaptive} expands the targeted-attack results summarized in the main paper. Local color smoothing preserves high bit accuracy across its measured severities. Opacity- and scale-guided purification reduce recovery at smaller visual losses than spatial partial theft, whose fragments have much larger ground-truth PSNR drops. These results separate attacks on the color readout from attacks that remove or alter its Gaussian support.

\subsection{Overwrite and copy collusion}

An independent-key overwrite re-embeds into an already marked model. Table~\ref{tab:overwrite} reports both the original owner's and the attacker's recovery after that second bounded solve. With smaller color budgets, the owner's message remains fully decodable while the attacker's recovery increases with its permitted change. At the same 0.02 bound used by the owner, both messages remain recoverable.

\begin{table}[!htbp]
    \caption{Targeted overwrite. Bit accuracy is evaluated for both the original owner and the attacker after the attacker's own bounded solve. Results average three scenes and three attacker keys.}
    \label{tab:overwrite}
    \centering
    \small
    \resizebox{0.76\textwidth}{!}{\begin{tabular}{@{}rrrrr@{}}
\toprule
Attacker bound & Cells & Owner bit acc. & Attacker bit acc. & Final/base PSNR \\
\midrule
0.005 & 9 & 100.00 & 59.72 & 40.47 \\
0.010 & 9 & 100.00 & 78.12 & 40.11 \\
0.020 & 9 & 95.49 & 100.00 & 38.10 \\
\bottomrule
\end{tabular}}
\end{table}
\FloatBarrier

Table~\ref{tab:collusion} evaluates per-Gaussian color averaging of differently keyed copies. A two-copy mixture retains 93.92\% member-key recovery while an unrelated key is near chance at 46.88\%; member-key recovery is 79.17\% for three copies. This test complements wrong-key calibration by probing keys that are actually present in a mixed asset.

\begin{table}[!htbp]
    \caption{Copy collusion by per-Gaussian linear-RGB averaging over Lego, Hotdog, and Bonsai. Member rows decode a key present in the mixture, and the non-member row tests an unrelated key.}
    \label{tab:collusion}
    \centering
    \small
    \resizebox{0.62\textwidth}{!}{\begin{tabular}{@{}llrrr@{}}
\toprule
Copies & Decoded key & Cells & Bit acc. & Colluded/base PSNR \\
\midrule
2 & non-member & 9 & 46.88 & 42.03 \\
2 & member & 18 & 93.92 & 42.03 \\
3 & member & 9 & 79.17 & 42.76 \\
\bottomrule
\end{tabular}}
\end{table}

\FloatBarrier
\section{Additional Viewpoint Fidelity}
\label{app:extended}

\subsection{Multi-shell views}

\begin{table}[h]
    \caption{Direct fidelity under multi-shell viewpoint sampling~\citep{matsubara2026msvs}. The view count follows the scene-specific two-shell sampling protocol.}
    \label{tab:msvs}
    \centering
    \small
    \resizebox{0.45\textwidth}{!}{\begin{tabular}{@{}lrrrr@{}}
\toprule
Scene & Views & PSNR & SSIM & LPIPS \\
\midrule
Bonsai & 37 & 38.74 & 0.988 & 0.0060 \\
Garden & 37 & 42.73 & 0.995 & 0.0015 \\
Chair & 134 & 44.35 & 0.998 & 0.0015 \\
Drums & 134 & 39.94 & 0.991 & 0.0028 \\
Ficus & 158 & 50.84 & 0.999 & 0.0002 \\
Hotdog & 126 & 42.32 & 0.994 & 0.0039 \\
Lego & 72 & 41.00 & 0.990 & 0.0029 \\
Materials & 121 & 42.60 & 0.993 & 0.0024 \\
Mic & 109 & 51.80 & 0.999 & 0.0002 \\
Ship & 109 & 40.94 & 0.993 & 0.0049 \\
\bottomrule
\end{tabular}}
\end{table}

Multi-shell evaluation samples 37 to 158 viewpoints per scene in the ten-scene cross-method corpus, including cameras away from the official test paths. Every sampled scene stays above 38.7\,dB direct PSNR and below 0.0060 LPIPS (Table~\ref{tab:msvs}), which supports the spatial reach of the fidelity result.

\FloatBarrier
\section{Carrier Geometry and Sensitivity Evidence}
\label{app:ablation}

\subsection{Carrier angular geometry}

\begin{figure}[!htbp]
    \centering
    \includegraphics[width=0.8\columnwidth]{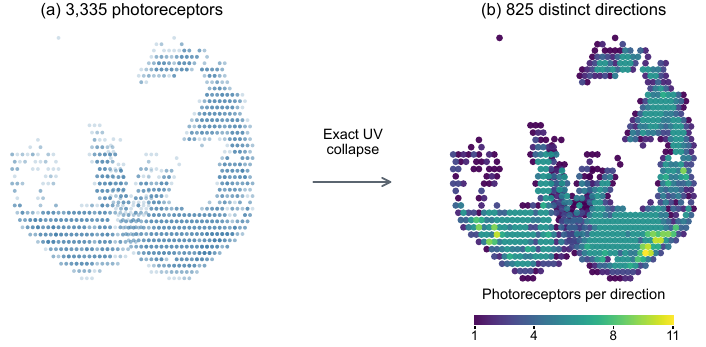}
    \caption{Connectome-derived carrier construction. Left: 3,335 mapped MaleCNS photoreceptors in the common display projection, with coincident samples overplotted. Right: 825 distinct projected coordinates after exact duplicate collapse. All right-panel markers have the same size; color shows the number of source photoreceptors mapped to each direction.}
    \label{fig:carrier}
\end{figure}

The source array is reconstructed from the published MaleCNS visual-system entries and matches all 3,335 expected photoreceptors exactly. Figure~\ref{fig:carrier} displays these samples in their common display projection and the 825 distinct coordinates obtained by exact duplicate collapse. Color encodes the number of photoreceptors at each retained direction. The resulting carrier is clustered and nonuniform; its construction is deterministic rather than scene-specific.

Table~\ref{tab:lattice_geometry} compares geometric summaries against three equal-count controls over the same forward cap. The connectome-derived directions cover 24.8\% of uniformly sampled cap directions at their local half-spacing cone radius, while the controls cover 58.7\% to 70.4\%.

\begin{table}[h]
    \caption{Angular geometry of the 825-direction lattices. Coverage is the fraction of 20,000 uniformly sampled cap directions within the method's cone radius. Random-cone values average the three registered seeds.}
    \label{tab:lattice_geometry}
    \centering
    \small
    \resizebox{0.79\textwidth}{!}{\begin{tabular}{@{}lrrrr@{}}
\toprule
Lattice & Mean NN angle & Cone radius & Covered cap & P95 nearest angle \\
\midrule
FlyMark & 1.037$^\circ$ & 0.532$^\circ$ & 24.80\% & 8.105$^\circ$ \\
Fibonacci & 1.770$^\circ$ & 0.881$^\circ$ & 70.41\% & 1.101$^\circ$ \\
Regular hex & 1.631$^\circ$ & 0.812$^\circ$ & 58.97\% & 1.434$^\circ$ \\
Random cone & 1.664$^\circ$ & 0.819$^\circ$ & 58.72\% & 1.201$^\circ$ \\
\bottomrule
\end{tabular}}
\end{table}

\subsection{One-factor sensitivity}

All sensitivity settings use the four development/validation scenes with key 17 and change one factor from the frozen reference, so the reference row has a different scope from the ten-scene main estimate. The grid traces a smooth capacity and fidelity trade-off: 16 bits are recovered perfectly in this subset and 64 bits retain 96.48\%; tighter color bounds raise PSNR and lower recovery, while a 0.04 bound reaches 100\% at 40.28\,dB; twelve or more orbit stops retain at least 97.66\% clean bit accuracy; and smaller cones improve both recovery and fidelity here, while larger cones increase overlap among the constraints of the joint solve.

\small
\begin{longtable}{p{2.5in}rrrr}
    \caption{One-factor-at-a-time sensitivity on four development/validation scenes. Bit accuracy and recomputed-mask accuracy are percentages.}\label{tab:sensitivity}\\
    \toprule
    Setting & Bit acc. & Recomputed & PSNR (dB) & LPIPS \\
    \midrule
    \endfirsthead
    \multicolumn{5}{c}{\tablename\ \thetable\ continued}\\
    \toprule
    Setting & Bit acc. & Recomputed & PSNR (dB) & LPIPS \\
    \midrule
    \endhead
    \bottomrule
    \endfoot
    Reference: $B=32$, $\epsilon=\Delta=0.02$, $T=24$ & 100.00 & 99.22 & 46.02 & 0.0009 \\
Message length = 16 & 100.00 & 100.00 & 46.05 & 0.0009 \\
Message length = 64 & 96.48 & 96.48 & 46.32 & 0.0009 \\
$\epsilon=0.005$ & 70.31 & 70.31 & 58.39 & 0.0000 \\
$\epsilon=0.01$ & 87.50 & 88.28 & 52.47 & 0.0002 \\
$\epsilon=0.04$ & 100.00 & 100.00 & 40.28 & 0.0032 \\
$\Delta=0.01$ & 100.00 & 100.00 & 47.20 & 0.0011 \\
$\Delta=0.04$ & 87.50 & 86.72 & 44.69 & 0.0013 \\
$\Delta=0.08$ & 77.34 & 78.12 & 43.52 & 0.0015 \\
Orbit stops = 8 & 94.53 & 93.75 & 48.65 & 0.0005 \\
Orbit stops = 12 & 97.66 & 97.66 & 47.82 & 0.0007 \\
Orbit stops = 48 & 98.44 & 99.22 & 45.01 & 0.0012 \\
Orbit radius = $1.0\rho$ & 100.00 & 100.00 & 47.01 & 0.0008 \\
Orbit radius = $1.25\rho$ & 100.00 & 100.00 & 46.46 & 0.0008 \\
Orbit radius = $2.0\rho$ & 92.19 & 92.19 & 47.12 & 0.0007 \\
Orbit phase = $3.75^\circ$ & 97.66 & 97.66 & 46.00 & 0.0010 \\
Orbit phase = $7.5^\circ$ & 98.44 & 98.44 & 46.13 & 0.0009 \\
Elevation = $0^\circ$ & 97.66 & 97.66 & 46.28 & 0.0009 \\
Elevation = $20^\circ$ & 99.22 & 99.22 & 46.26 & 0.0009 \\
Elevation = $30^\circ$ & 97.66 & 97.66 & 45.98 & 0.0010 \\
Cone multiplier = 0.25 & 100.00 & 100.00 & 48.32 & 0.0006 \\
Cone multiplier = 0.75 & 89.06 & 88.28 & 45.66 & 0.0009 \\
Cone multiplier = 1.0 & 88.28 & 85.94 & 46.19 & 0.0007 \\
\end{longtable}
\normalsize

\section{Two-Dimensional Image Extension}
\label{app:2d}

We test whether the retinal carrier and keyed QIM code also work with raster
images on the official BSDS500 natural-image collection~\citep{arbelaez2011contour}.
All 100 validation images select the image-domain QIM step; all 200 disjoint
test images are evaluated once with each of three keys. Images retain their
native $321\times481$ or $481\times321$ pixel dimensions. There is no
image-specific training or learned decoder.

Let $X_p^{\mathrm{lin}}\in[0,1]^3$ be linear RGB at pixel $p$, and let
$A^{\mathrm{img}}_{jp}$ be the bilinear sampling weight at retinal UV location
$j$ after exact collapse to 825 unique locations. The image-domain reading is
$r_j=\sum_p A^{\mathrm{img}}_{jp}\,\boldsymbol{w}^{\top}X_p^{\mathrm{lin}}$,
where $\boldsymbol{w}=(0.2126,0.7152,0.0722)^{\top}$ is the luminance vector.
Eligible readings lie in $[0.05,0.95]$. As in the 3DGS method, keyed QIM
assigns eligible readings to message bits modulo the message length and sets
parity-constrained targets $q_j$. An achromatic scalar change $\delta_p$ to
each writable pixel is found in one sparse bounded least-squares solve:
\begin{equation}
\min_{\boldsymbol{\delta}}\sum_{j\in E}
\left(\sum_p A^{\mathrm{img}}_{jp}\delta_p-(q_j-r_j)\right)^2,
\quad |\delta_p|\leq0.02,\quad
0\leq X_{p,c}^{\mathrm{lin}}+\delta_p\leq1\ \text{for every }c.
\end{equation}
The result is stored as an 8-bit sRGB PNG. Quantized channel values are
restricted to those that preserve the same $0.02$ linear-RGB bound in the
stored image, which is then reopened for extraction. The original eligible
sample set is retained in the verification record.

We evaluate QIM steps $\{0.008,0.012,0.016\}$ with key 17 on the complete
validation split, ranking by clean bit accuracy and breaking ties by direct
PSNR. This selects $0.008$ before evaluating the test split.
Each test image is marked with keys 17, 29, and 43 and measured after PNG
storage. The primary metric is correct message bits divided by $B$; direct
PSNR compares the saved PNG with the decoded source image in sRGB. We average
keys within each image and then average images; intervals resample the 200
images 10,000 times. For ownership controls, the
unmarked image is read under a matching key, while each wrong candidate key
independently determines its dither and expected message.

Table~\ref{tab:2d} reports bit accuracy after PNG storage for the marked
images and both ownership controls.

\begin{table}[h]
    \caption{2D image bit accuracy on the complete BSDS500 test split (200 images, three marking keys per image). Higher is better. Confidence intervals use image-first bootstrap resampling; wrong-key scores average three candidate keys per marked image.}
    \label{tab:2d}
    \centering
    \small
    \begin{tabular}{@{}lrr@{}}
        \toprule
        Condition & Bit accuracy (\%) & 95\% CI (\%) \\
        \midrule
        Marked PNG & 99.95 & 99.88 to 100.00 \\
        Unmarked, matched key & 49.15 & 48.43 to 49.85 \\
        Marked, wrong candidate key & 50.50 & 50.37 to 50.63 \\
        \bottomrule
    \end{tabular}
\end{table}

\FloatBarrier

\section{Additional Fixed-View Qualitative Results}
\label{app:qualitative}

\begin{figure}[h!]
    \centering
    \includegraphics[width=\textwidth]{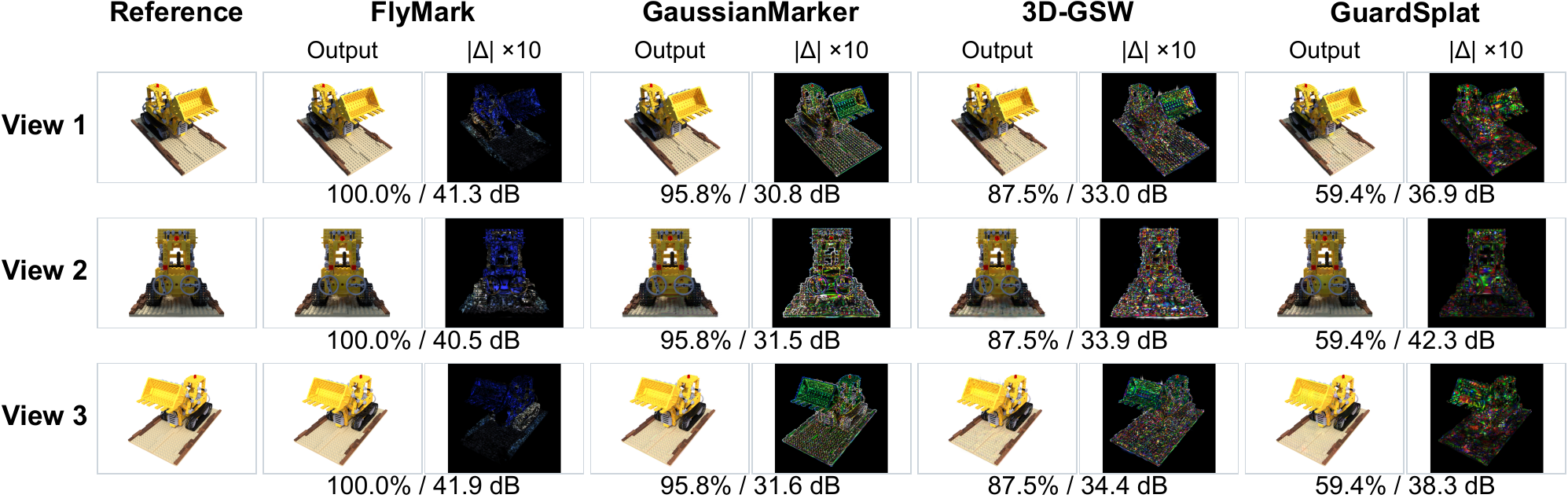}
    \caption{Lego at registered frames 40, 100, and 159. For \method, GaussianMarker, 3D-GSW, and GuardSplat, the equal-size panel to the right of each output shows its $10\times$ absolute RGB difference from the matched reference. The one-line values are key-17 bit accuracy (\%) / direct PSNR (dB), with PSNR evaluated at the displayed frame.}
    \label{fig:qual_lego}
\end{figure}

\begin{figure}[h!]
    \centering
    \includegraphics[width=\textwidth]{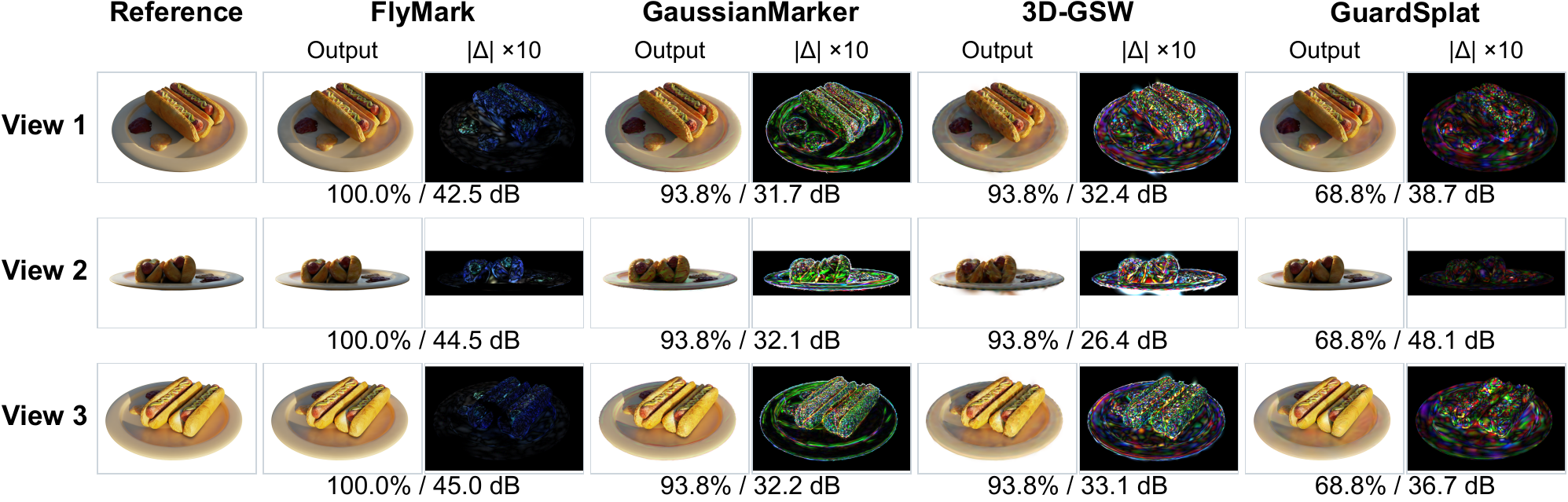}
    \caption{Hotdog at registered frames 40, 100, and 159. For \method, GaussianMarker, 3D-GSW, and GuardSplat, the equal-size panel to the right of each output shows its $10\times$ absolute RGB difference from the matched reference. The one-line values are key-17 bit accuracy (\%) / direct PSNR (dB), with PSNR evaluated at the displayed frame.}
    \label{fig:qual_hotdog}
\end{figure}

\begin{figure}[h!]
    \centering
    \includegraphics[width=\textwidth]{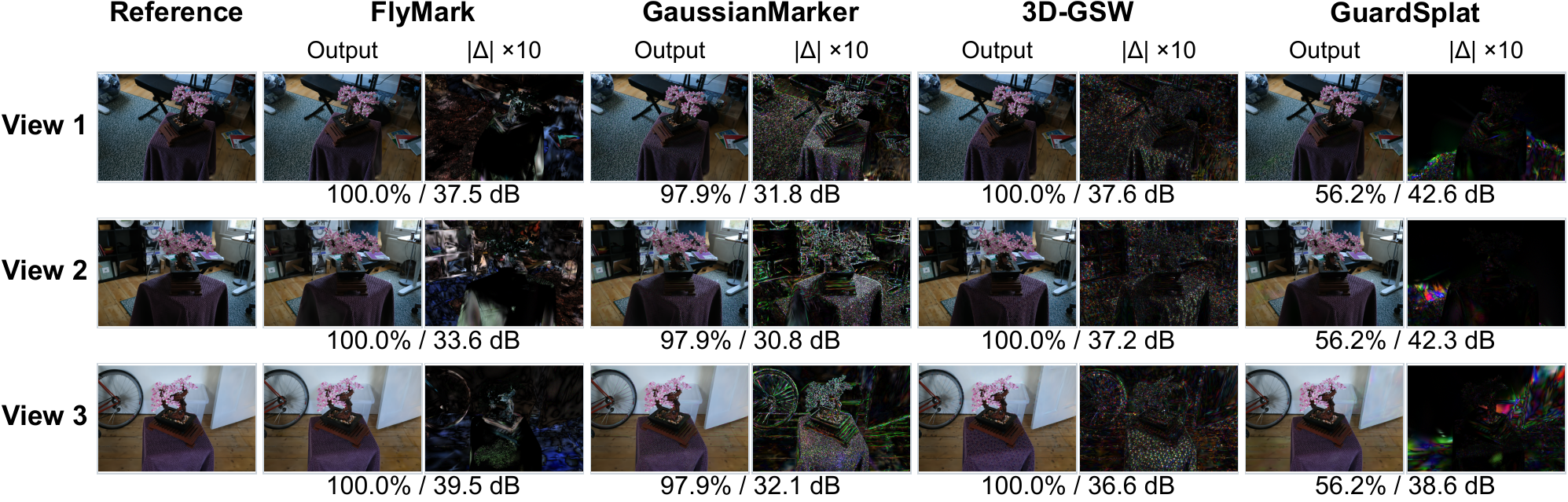}
    \caption{Bonsai at registered frames 7, 18, and 29, including the targeted GuardSplat key-17 qualitative cell. For every method, the equal-size panel to the right of each output shows its $10\times$ absolute RGB difference from the matched reference. The one-line values are key-17 bit accuracy (\%) / direct PSNR (dB), with PSNR evaluated at the displayed frame.}
    \label{fig:qual_bonsai}
\end{figure}

Figures~\ref{fig:qual_lego}, \ref{fig:qual_hotdog}, and~\ref{fig:qual_bonsai} contain every registered qualitative view, taken at normalized positions 0.2, 0.5, and 0.8 of the sorted official test split. The indices were fixed before metric inspection, and each row compares all four published methods at exactly the same camera and crop. Every marked output is followed immediately by an equal-size absolute RGB difference from the matched reference, multiplied by ten under a shared display transform. In the line below each pair, the value before the slash is key-17 bit accuracy (\%) and the value after it is direct PSNR (dB) for that frame; aggregate results are reported in Tables~\ref{tab:main} and~\ref{tab:roles}.

\end{document}